\documentclass{article}
\usepackage{spconf}
\usepackage[fleqn]{amsmath}
\usepackage{graphicx}
\usepackage{times}
\usepackage{epsfig}
\usepackage{graphicx}
\usepackage{amsmath}
\usepackage{amssymb}
\usepackage{supertabular}
\usepackage{graphicx,subfigure}
\usepackage{multirow}
\usepackage[bottom]{footmisc}
\usepackage{url}
\usepackage{color}
\usepackage{subfigure}
\usepackage{float}
\usepackage{spconf}
\usepackage{textcomp}
\usepackage{gensymb}

\begin{document}
%
\title{Neural Networks with Manifold Learning for Diabetic Retinopathy Detection}
\makeatletter
\def\@name{ \emph{Arjun Raj Rajanna}, \emph{Kamelia Aryafar*},\hspace{1mm}\emph{ Rajeev Ramchandran**},  \\ \emph{Christye Sisson}, \emph{ Ali Shokoufandeh*} , \emph{ Raymond Ptucha} }
\makeatother

\address{Rochester Institute of Technology, * Drexel University, **University of Rochester}

%
%
%

\maketitle

\begin{abstract}
 Widespread surveillance programs using remote retinal imaging has proven to decrease the risk from diabetic retinopathy, the leading cause of blindness in the US. However, this process still requires manual verification of image quality and grading of images for level of disease by a trained human grader and will continue to be limited by the lack of such scarce resources. 
Computer-aided diagnosis of retinal images have recently gained increasing attention in the machine learning community. In this paper, we introduce a set of neural networks for diabetic retinopathy classification of fundus retinal images. We evaluate the efficiency of the proposed classifiers in combination with preprocessing and augmentation steps on a sample dataset. Our experimental results show that neural networks in combination with preprocessing on the images can boost the classification accuracy on this dataset. Moreover the proposed models are scalable and can be used in large scale datasets for diabetic retinopathy detection. The models introduced in this paper can be used to facilitate the diagnosis and speed up the detection process.
\end{abstract}

\maketitle

\section{Introduction}

Diabetic retinopathy (DR) is a severe and common eye disease
which is caused by changes in the retinal blood vessels. Diabetic retinopathy is the leading cause of blindness in the working age US population (age 20-74 years). Ninety-five percent of this vision loss is preventable with timely diagnosis. The
national eye institute~\footnote{https://nei.nih.gov/eyedata/diabetic} estimates that from 2010 to 2050,
the number of Americans with diabetic retinopathy is expected
to nearly double, from 7.7 million to 14.6 million. By 2030, a half a billion people worldwide are expected to develop diabetes and a third of these likely will have DR. DR can
be detected during a dilated eye exam by an ophthalmologist
or optometrist in which the pupils are pharmacologically dilated and the retina is examined with specialized condensing lenses and a light source in real time. Primary barriers for diabetic patients in accessing timely eye care include lack of awareness of needing eye exams, inequitable distribution of eye care providers and the addition costs in terms of travel, time off from work, and additional medical fees of seeing another care provider~\cite{america2002vision}. Tele-ophthalmology, or using non-mydriatic camera in non-ophthalmic settings, such as primary care offices, to capture digital images of the central region of the retina in diabetic patients has increased the rate of annual diabetic retinopathy detection~\cite{zhang2008measuring}. This methodology has been validated against dilated retinal examinations by eye specialists and is accepted by the UK National Health Service and US National Center for Quality Assurance. Incorporating an individual’s retinal photos for those with diabetic retinopathy has been shown to be associated with better glycemic control for poorly controlled diabetic patients~\cite{liew2014comparison}.

\begin{figure}[t!]
        \centering
        \subfigure[]{
                \includegraphics[scale =0.15] {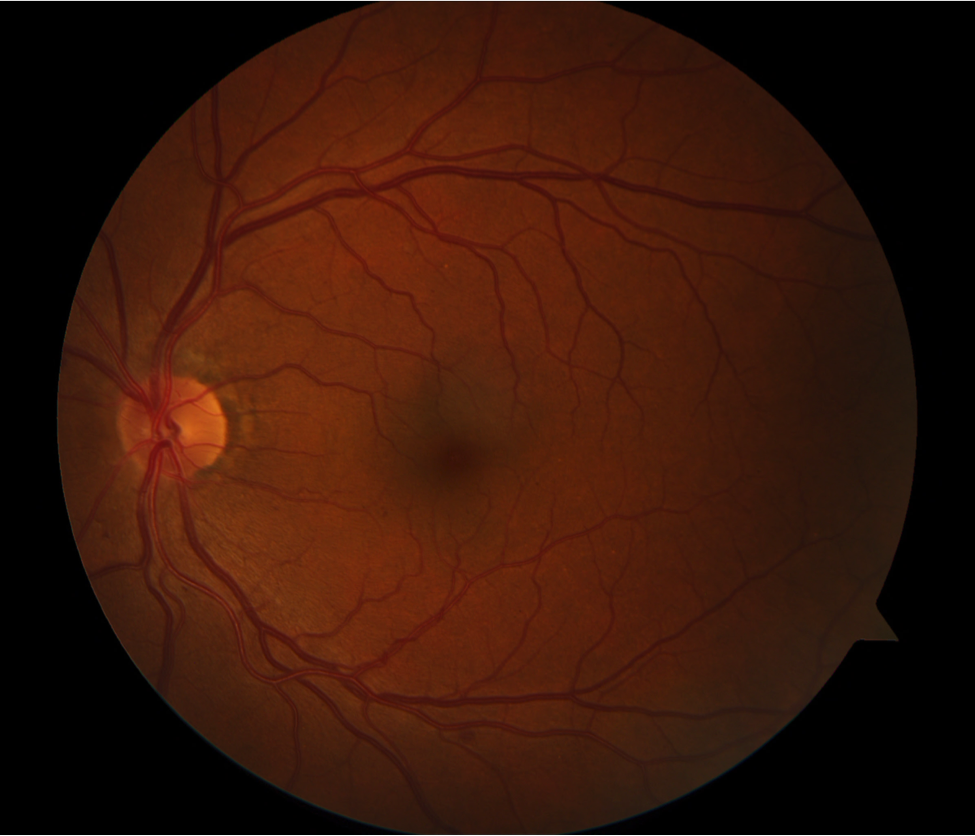}
       }
        \subfigure[]{
                \includegraphics[scale =0.148] {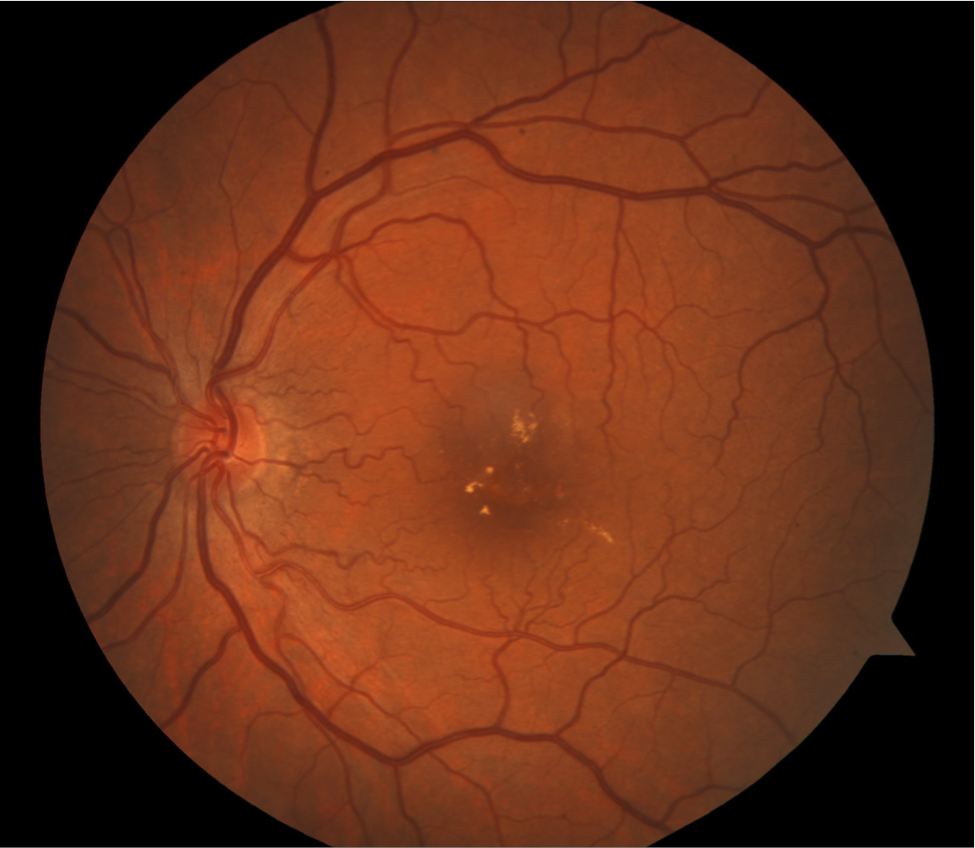}
       }
        \subfigure[]{
                \includegraphics[scale =0.202] {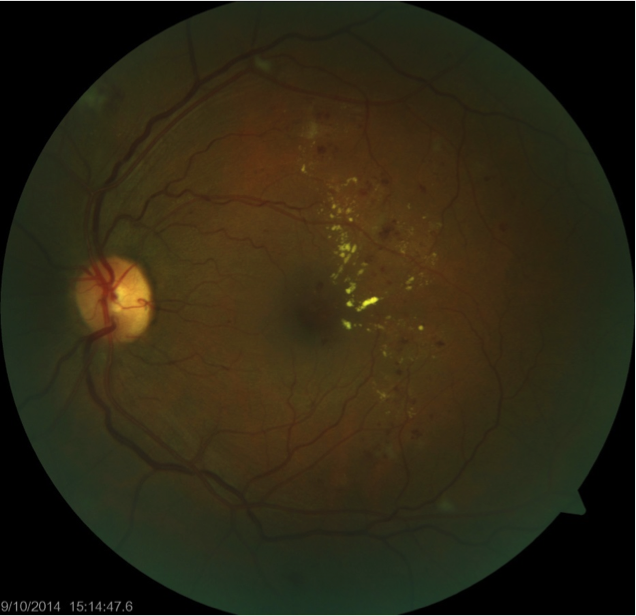}
       }
        \subfigure[]{
                \includegraphics[scale =0.24] {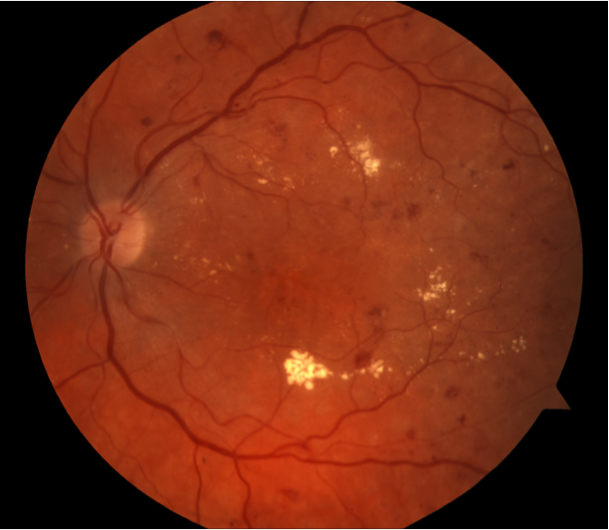}
       }
        \subfigure[]{
                \includegraphics[scale =0.20] {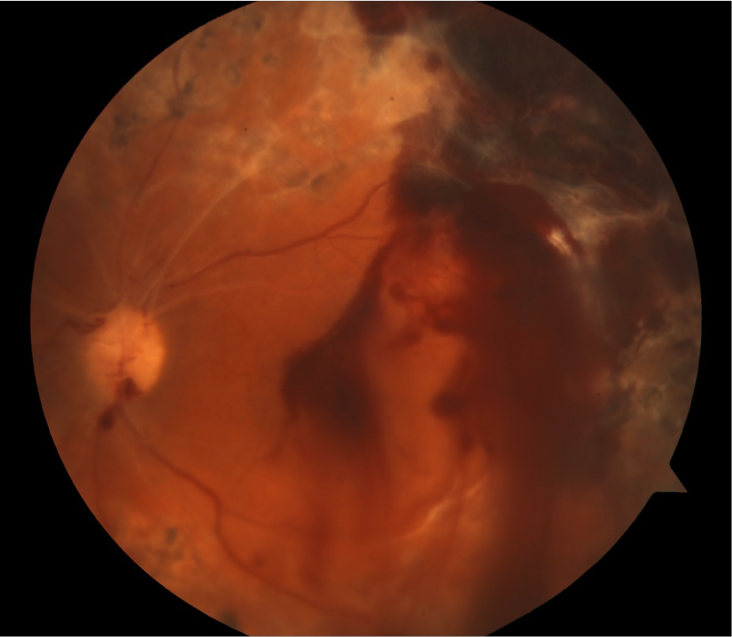}
       }
         \label{fig:sample}
         \centering
        \caption{Illustration of the stages of DR development: (a) Normal , (b) Mild, (c) Moderate ,
          (d) Severe  and (e) Proliferative diabetic retinopathy conditions. }
\end{figure}
 
The procedures for DR detection
can often be difficult and dependent on scarce resources such as
trained medical professionals. The severity of this wide-spread
problem then in combination with the cost and efficiency of
the detection procedures motivates computer-aided models that
can learn the underlying patterns of this disease and speedup
the detection process. In this paper, we introduce a set of
neural networks (NNs) that can facilitate diabetic retinopathy
detection by training a model on color fundus photography
inputs.

 \begin{figure*}[t!]
         \centering
         \subfigure[]{
                 \includegraphics[scale =0.15] {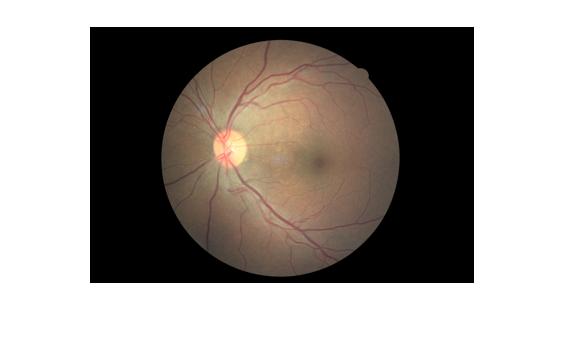}
        }
         \subfigure[]{
                 \includegraphics[scale =0.15] {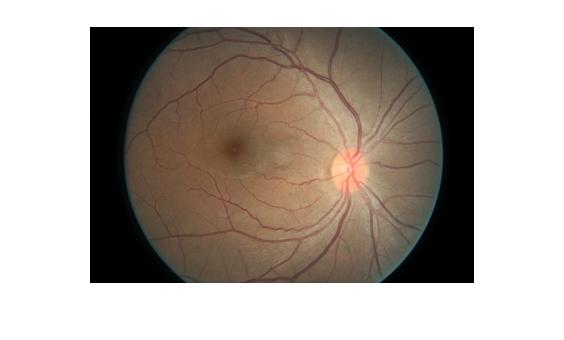}
        }
         \subfigure[]{
                 \includegraphics[scale =0.15] {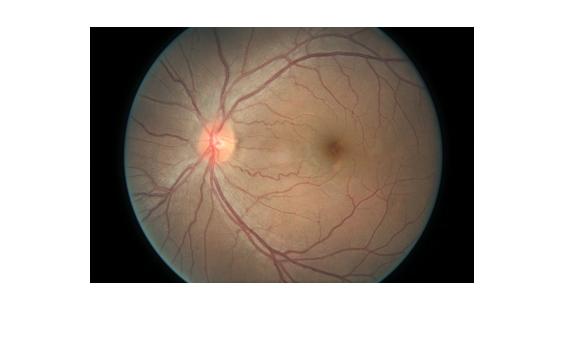}
        }
         \subfigure[]{
                 \includegraphics[scale =0.15] {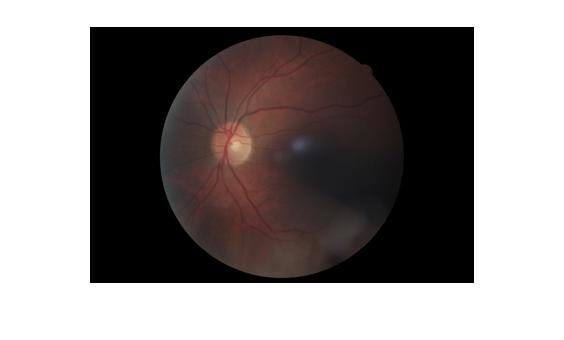}
        }
         \subfigure[]{
                 \includegraphics[scale =0.15] {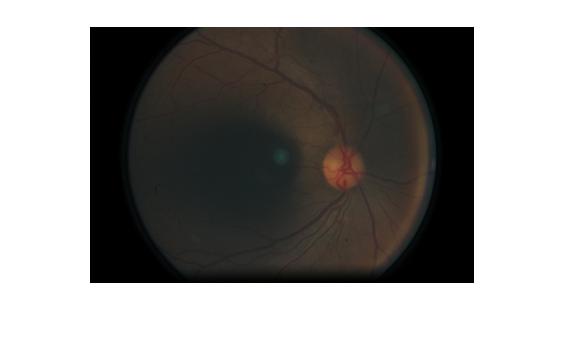}
        }
          \label{fig:sample2}
          \centering
         \caption{Input images from the sample dataset in five labeled classes: (a) Normal , (b) Mild, (c) Moderate , (d) Severe  and (e) Proliferative DR are shown. }
                   \label{fig:sample2}
 \end{figure*}

Fundus photography or fundography is the process of photographing of the interior surface of the eye, including the retina, optic disc, macula, and posterior pole ~\cite{lee2000chin}. Fundus photography is used by trained medical professionals for monitoring progression or diagnosis of a disease. Figure~\ref{fig:sample} illustrates example fundography images from a sample dataset for various stages ~\footnote{The stages of DR correspond to class labels for DR detection and classification. Class labels are assigned based on dividing the stages of disease into 5 different classes as illustrated in Figure~\ref{fig:sample2}.} of diabetic retinopathy disease.

The digital fundus images often contain a set of lesions which can indicate the signs of DR. Automated DR detection algorithms are composed of multiple modules: image preprocessing, feature extraction and model fitting (training). Image preprocessing reduces noise and disturbing imaging artifacts, enhances image quality using appropriate filters and defines the region of interest (ROI) within the images~\cite{cree1999preprocessing}. Illumination and quality variations often make preprocessing an important step for DR detection. Several attempts in computer vision literature have been proposed to best represent the ROI~\cite{niemeijer2005automatic,walter2002contribution} in fundus images. Image enhancement, mass screening (including detection of pathologies and retinal features), and monitoring (including feature detection and registration of retinal images) are often cited as three major contributions of preprocessing to DR~\cite{walter2002contribution}. Green-plane representation of RGB fundus images to increase the contrast for red lesions~\cite{niemeijer2005automatic} and shadow correction to address the background vignetting effect~\cite{hoover2003locating} are perhaps among the most popular preprocessing steps for the task in hand~\cite{spencer1996image,frame1998comparison}. Landmark and feature detection~\cite{gardner1996automatic} and segmentation~\cite{staal2004ridge,spencer1996image} on digital fundus images have also been explored as preprocessing steps.

The second step in developing an automatic DR detection algorithm is the choice of a learning model or classifier. DR detection has been explored through various classifiers such as fuzzy C-means clustering~\cite{sopharak2009automatic}, neural networks ~\cite{osareh2001automatic} and a set of feature vectors on benchmark datasets~\cite{walter2002contribution}. While the detection and preprocessing algorithms appear to be maturing, the automatic detection of DR remains an active area of research. Scalable detection algorithms are often required to validate the results on larger, well-defined, but more diverse datasets. The quality of input fundus photographs, image clarity problem, lack of available  benchmark datasets and noisy dataset labels often present challenges in automating the diabetic retinopathy detection of input data or developing a robust set of feature vectors to represent the input data. In this paper, we explore neural networks with domain knowledge preprocessing and introduce manifold learning techniques that can boost the classification accuracy on a sample dataset.


This paper is organized as follows: In section~\ref{sec:method} we present the outline of our proposed model. Preprocessing is presented in Section~\ref{sec:pre}. We present the comparative results of DR detection on a sample dataset in Section~\ref{sec:results}. Finally we conclude this paper in Section~\ref{sec:conclusion} and propose future research directions.

\section{Method}
   \label{sec:method}   

The input data to our DR detection model is a set of RGB represented fundus images. The proposed method involves four successive sequences: (i) preprocessing, (ii) data augmentation, (iii) manifold learning and dimensionality reduction techniques, and, (iv) neural network design. In this section we describe each step in details.


       

       \subsection{Data Preprocessing}
       \label{sec:pre}
 The original digital fundus images in the sample dataset are high resolution RGB images with different sizes. The goal of the preprocessing step is contrast enhancement, smoothing and unifying images (resizing to same pixel resolution across images with minimal ROI loss) and enhancement of the red lesion representation.

The first step in preprocessing the images is resizing all training and testing images to $256\times388$ pixel resolution. Once the original images are resized, we then represent the resized images in green channel only. Green channel representation of resized fundus images is considered as the natural basis for
vessel segmentation since it normally presents a higher
contrast between vessels and retinal background. Using only the green channel also replicates a common image practice in ophthalmology of using a green filter to capture the retinal image. This image, called \textit{red-free}, is commonly used to increase local contrast to help identify hemorrhage, exudate or other retinal disease signs. Moreover, switching the RGB image to its green channel form also reduces the data dimensionality to one third of the original RGB presentation and reduces the time complexity of the detection algorithm.

After green channel representation of resized fundus images, we apply a histogram equalization on the green channel to adjust image intensities and enhance the contrast around the red lesion. Figure~\ref{fig:imagecondition} shows two different images from class 3 (severe) and class 4 (proliferative) stages of DR with green channel representation before and after histogram equalization.

\begin{figure}[t!]
        \centering
        \subfigure[]{
                \includegraphics[scale =0.125] {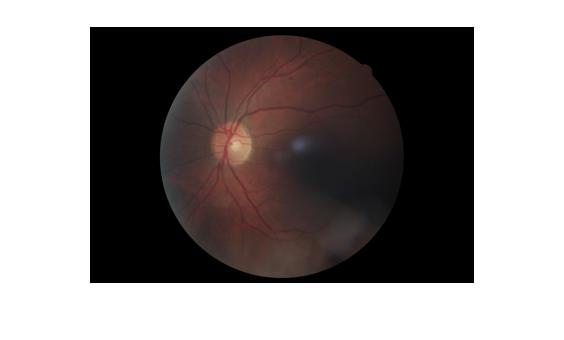}
                \label{fig:a}
       }
        \subfigure[]{
                \includegraphics[scale =0.125] {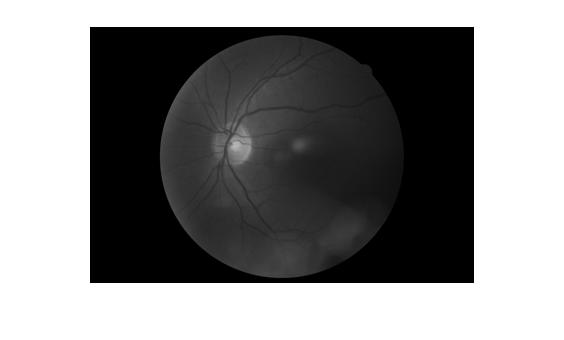}
                \label{fig:b}
       }
        \subfigure[]{
                \includegraphics[scale =0.125] {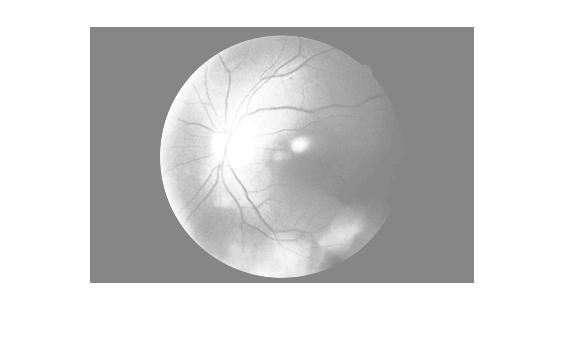}
                \label{fig:c}
       }
       \subfigure[]{
                   \includegraphics[scale =0.125] {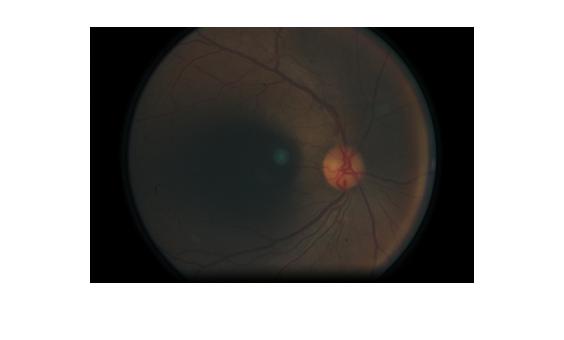}
                \label{fig:d}
       }
        \subfigure[]{
                \includegraphics[scale =0.125] {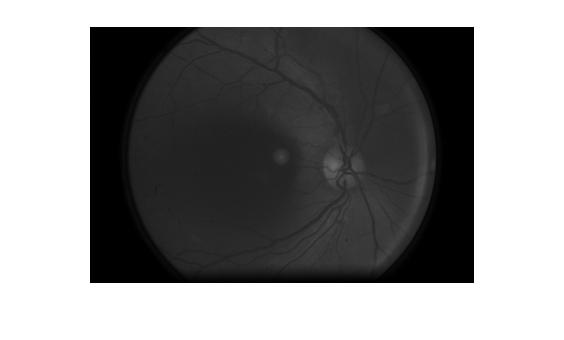}
                \label{fig:e}
       } 
       \subfigure[]{
                \includegraphics[scale =0.125] {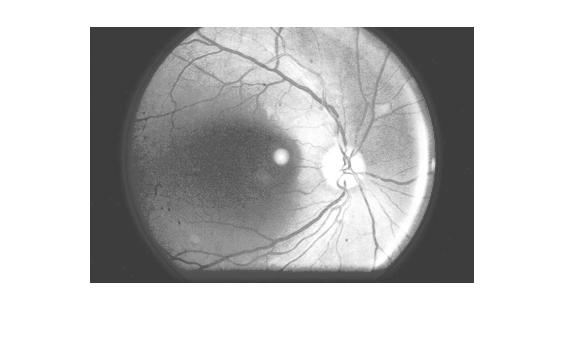}
                \label{fig:f}
       }
      
         \label{fig:imagecondition}
         \centering
        \caption{ Class 3 and class 4 samples: (a) Original image , (b) Green channel (Red-free) and  (c) Green channel histogram equalization for an example of class 3 (severe DR) are illustrated. (d) Original image , (e) Green channel (Red-free) and  (f) Green channel histogram equalization for an example of class 4 (proliferative DR) are illustrated.}
                 \label{fig:imagecondition}
\end{figure}

  \subsection{Data Augmentation}
  \label{sec:aug}

\begin{table}[h!]
   \label{tab:samples}
   \caption{Sample Dataset Breakdown}
    \begin{center}
\begin{tabular}{|c||c||c|} 
\hline
DR Class & Original Sample Size & Augmented Size\\
\hline
 \hline
0-Normal& $100$ & N/A\\ \hline
1-Mild& $59$ & $100$\\ \hline
2-Moderate&$148$ & $20$\\ \hline
3-Severe&$28$ & $100$\\ \hline
4-proliferative&$26$ & $100$\\ \hline
 \hline
 \end{tabular}
   \end{center}
    \label{tab:samples}
\end{table}

   In the proposed method we use example duplication and rotation as the augmentation method. The input images have been duplicated within classification folds for under-represented classes~\footnote{The data has been duplicated within folds to avoid classification on samples appearing on both testing and training splits of the classifie.}. We rotate images with angles of $45\degree$, $90\degree$, $135\degree$, $180\degree$, $225\degree$ and $270\degree$ degrees. Finally the augmented dataset is randomly sampled among all rotated preprocessed images. Table~\ref{tab:samples} summarizes the dimensionality of each sample size before and after augmentation for all different classes. We note that the $0$-class (normal DR stage label) has been excluded from data augmentation due to large number of examples in the original dataset.
 
\subsection{Manifold Learning}
\label{sec:manifold}
Manifold learning is a non-linear dimensionality reduction technique that is often applied in large scale machine learning models to address the curse of dimensionality~\cite{lee2007nonlinear} and reduce the time complexity of the learning models. In this approach, it is often assumed that the original high-dimensional data lies on an embedded non-linear low dimensional manifold within the higher-dimensional space. 

Let matrix $M \in R^{N \times k}$ represent the input matrix of all preprocessed examples in our dataset, where each row represents an example unrolled preprocessed image vector. $N$ represents the total number of input images and $k$ is the dimensionality of the unrolled image representation in green channel with histogram equalization. If the attribute or representation dimensionality $k$ is large, then the space of unique possible rows is exponentially large and sampling the space becomes difficult and the machine learning models can suffer from very high time complexity. This motivates a set of dimensionality reduction and manifold learning techniques to map the original data points in a lower-dimensional space. In this paper, we use principal component analysis (PCA), supervised locality preserving projections (SLPP)~\cite{shan2005appearance}, neighborhood preserving embedding (NPE)~\cite{he2005neighborhood} and spectral regression (SR)~\cite{cai2007spectral} as dimensionality reduction and manifold learning techniques.

\subsection{Neural Networks Architecture}    
\label{sec:model}
The input to our neural network is the preprocessed unrolled image data after dimensionality reduction steps. All networks have been constructed with 2 layers and different number of neurons (160 and 100). The input data matrix $M \in R^{N \times k}$ of unrolled preprocessed fundus images are first normalized by mean subtraction and division by standard deviation. The Rectified Linear Units (ReLUs), $\phi(x) = max(0,x)$, has been selected as the activation function to provide the network with invariance to scaling, thresholding above zero, faster learning and non-saturating for large-scale data~\cite{dahl2013improving}. 


\begin{table}[t!]
   \label{tab:results}
   \caption{Classification accuracy \textbf{($\%$)} on sample dataset: }
    \begin{center}
\begin{tabular}{|c||c||c||c||c|} 
\hline
Prep&Aug&DR-method&Model&ACC($\%$)\\
\hline
 \hline
GRAY& N & PCA & 2-NNs & $66.43\%$\\ 
\hline
\hline
GRH& Y & PCA & NNs(160)&$82.24\%$\\ \hline
GRH& Y & SLPP & NNs(160)&$82.02\%$\\ \hline
GRH& Y & NPE & NNs(160)&$\mathbf{84.22}\%$\\ \hline
GRH& Y & SR & NNs(160)&$79.23\%$\\ \hline
GRH& Y & PCA & $k$-NN ($k=5)$ &$\mathbf{34.07}\%$\\ \hline
\hline
\hline
GRH& N & PCA & NNs(100) &$64.24\%$\\ \hline
GRH& N & SLPP & NNs(100) &$58.25\%$\\ \hline
GRH& N & SR & NNs(100) &$58.64\%$\\ \hline
GRH& N & NPE & NNs(100) &$64.94\%$\\ \hline
 \end{tabular}
   \end{center}
    \label{tab:results}
\end{table}

In this paper, we adopt the cross entropy as the similarity measure to penalize wrong guesses for labels and avoiding convergence to local minima in the optimization. The cross entropy is often described as:
$$\sum \limits_{i} \log( \frac{e^{x}}{\sum \limits_{i}^{ } e^{x}}).$$ Next, we select the regularization of the weights as an integral part of calculating the overall cost function. We adopt an $\ell_2$-regularization with penalty $\lambda$, where $\lambda$ is set to $1e-5$ experimentally. Finally, the overall cost function is the sum of cross entropy and weight penalty as:
$$- \sum \limits_{i}^{ } log( \frac{e^{x}}{\sum \limits_{i}^{ } e^{x}})  +  C\lambda \sum |w|^2.$$
We the back propagate to learn the weights. A variation of the BFGS algorithm, namely the L-BFGS is used in our model to estimate the parameters of the learning model~\cite{liu1989limited}. The L-BFGS does an unconstrained minimization of smooth functions using a linear memory requirement. 

\section{Experimental Results}  

\label{sec:results}
In this section we explain the details of our experiments on a public sample dataset for a recent Kaggle competition~\footnote{https://www.kaggle.com/c/diabetic-retinopathy-detection}. The dataset is subsampled to include around 1000 fundus photographs of diabetic retinopathy in five different stages: normal, mild, moderate, severe and proliferative labeled by trained medical professionals as illustrated in Figure~\ref{fig:sample2}~\footnote{It should be noted that the small scale sample dataset has been used in this experiments to speed up the results and not interfere with the competition results. The authors note that these models are scalable and can be repeated on large scale datasets as well.}. 

Table~\ref{tab:samples} shows the original sample size break-down by class before (second column) and after (third column) data augmentation. All input images have been resized to $256\times 388$ pixels and all experiments are performing a 5-fold cross validation. Gray scale representation of resized fundus images without additional preprocessing and augmentation has been used as the baseline dataset to highlight the contributions of data preprocessing. A baseline classifier, $k$-nearest neighbors ($k$-NN) has also been tested against the neural network architectures to emphasize the contributions of the neural networks in DR detection. We report the average classification accuracy rate for the baseline model and neural networks across all experiments. Table~\ref{tab:results} summarizes these results.

The first round of experiments evaluates the performance of neural network on gray scale fundus photographs. The gray-scale representation of these images have been provided as the input to a 2 layer neural network with $160$ neurons. It should be noted that there is no dimensionality reduction technique present in this experiment. As shown in the second row of Table~\ref{tab:results}, the average classification accuracy rate for this experiment is $66.43\%$.

Next, we experiment with preprocessed fundus images, where images are presented as green channel with histogram equalization (GRH). Rows 3 to 7 shows the classification results on this dataset with preprocessing and augmentation and different dimensionality reduction techniques. The neural networks are all 160 neurons and the baseline $k$-NN classifier uses $k=5$. The NPE reduces the input dimensionality to $154 \pm 3$ and is achieving the highest classification accuracy on this dataset. PCA with $161$, SLPP with $155 \pm 2$ and SR with $4$ as the reduced dimensionality are all outperforming the $k$-NN with the lowest classification accuracy rate across all experiments.

Finally, we repeat the experiments with preprocessed images and no augmentation to highlight the marginal effects of augmentation. Table~\ref{tab:results} rows 8 to 11 shows the result of these experiments with different dimensionality reductions for a 2 layer neural networks. The neural networks are all designed with $100$ neurons. PCA, SLPP, NPE all reduce the input dimensionality to $256$ and SR reduces this to $4$. Our experimental results show that, without data augmentation to enhance the robustness to variations in input and imbalanced classes, neural network with NPE is achieving the highest classification accuracy rate in this group (without augmentation).

\section{Conclusion and Future Work}
\label{sec:conclusion}
Diabetic retinopathy detection has been gaining increasing attention in computer vision and machine learning communities. The advanced digital retinal photography known as fundus photographs have often be used as the input for feature extraction, disease classification and detection. In this paper we first introduced a set of preprocessing and data augmentation techniques that can be utilized to enhance the image quality and data presentation in fundus photographs. we explored green channel and histogram equalization to represent the fundus photographs in a more compact manner for a neural network classifier.

Next, we explored a family of manifold learning techniques in conjunction with neural networks. Our experimental results showed that a neural network classifier with manifold learning and a more revealing preprocessing step can enhance the classification accuracy while providing a scalable model for diabetic retinopathy detection. 

In future, we plan to apply the recently popular convolutional neural networks (CNNs)~\cite{krizhevsky2012imagenet}, which has been shown effective for computer vision tasks, on this dataset. We anticipate that CNNs can boost the DR detection accuracy at scale.





\bibliographystyle{IEEEbib}
\bibliography{ref}

\end{document}